# In Vivo Quantification of Clot Formation in Extracorporeal Circuits

Rabin GERRAH [a,1] and Omid DAVID [b]
[a] *Department of Surgery, Division of Pediatric Cardiac Surgery, Oregon Health and Science University, Doernbecher Children's Hospital, Portland, OR*
[b] *Department of Computer Science, Bar-Ilan University, Ramat-Gan 52900, Israel*

**Abstract.** Clot formation is a common complication in extracorporeal circuits. In this paper we describe a novel method for clot formation analysis using image processing. We assembled a closed extracorporeal circuit and circulated blood at varying speeds. Blood filters were placed in downstream of the flow, and clotting agents were added to the circuit. Digital images of the filter were subsequently taken, and image analysis was applied to calculate the density of the clot. Our results show a significant correlation between the cumulative size of the clots, the density measure of the clot based on image analysis, and flow duration in the system.

## Introduction

Blood clots are formed as blood is exposed to any artificial surface. These surfaces include essential life supporting devices such as hemodialysis circuit or extracorporeal membrane oxygenator (ECMO) [1]. To prevent the formation of clots, a blood thinner is commonly applied. The proper clinical management of a patient with any extracorporeal circuit is based on finding the appropriate balance between the thrombosis from activation of clotting cascade due to exposure to a foreign material, and excessive bleeding due to blood thinning medications. Nevertheless, bleeding and thrombosis are the most important major morbidities during any type of extracorporeal circulation.

Various techniques have been developed to prevent the formation of blood clots due to foreign material exposure, and thus, eliminate the need for blood thinners. However, currently there is no available method for quantifying the magnitude of clot formation in the circuits. Current systems assess the thrombotic status indirectly, e.g., by measuring the levels of clotting factors, or alternatively by mechanically measuring the pressure gradient across a barrier as an indicator for system blockage due to formation of clots [2–4].

Currently, no method is available for evaluating the real-time clot formation in the circuit. A system monitoring and measuring the clot formation in real-time would be highly valuable in clinical settings for immediate identification of clots in the circuit, or for comparison and quantification of the amount of clots whenever a modality to reduce clot formation is assessed.

---

[1] Corresponding Author. Email: gerrah@ohsu.edu.

Furthermore, quantification of the clot formation in the clinical or research settings will facilitate studying and comparing various measures administered in order to minimize the amount of clots.

## 1. Method and Results

In this section we describe a novel image processing based method for measurement and quantification of blood clots.

We assembled a closed extracorporeal circuit including an oxygenator, and circulated blood at varying speeds, flowing in the range of 200–500 ml/min. A manifold of blood filters were placed in downstream of the flow. Clotting agents were added to the circuit, and the blood was left to start the coagulation process. Pressure was monitored across each filter to assess the extent of filter occlusion with clots. The circulation was discontinued in each branch after a fixed time interval had elapsed. Each filter was subsequently irrigated with normal saline, and a digital image of the filter was taken (Figure 1).

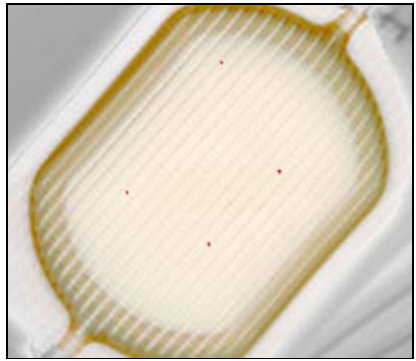

**Figure 1.** Blood filter.

We then analyzed the obtained images using ImageJ software developed by NIH[2]. After calculating the density of the clot, we observed that the larger the clot was on the filter, the higher the density of the clot pixels was in the image.

Having established the benefit of image analysis for quantification of clot formation, we developed a software for automatic quantification of clot density. After performing preprocessing (Figure 2), the software applies a binary filter to convert the image to binary format, where each pixel is either white or black (Figure 3), and then detects the clots by applying *connected component labeling* algorithm [5,6], where each blood clot is detected as a separate component (Figure 4). The number of members (pixels) in each detected component corresponds to the density of the clot.

Our results show a significant correlation between the size of the clots, pixel density and flow duration in the system.

---

[2] http://rsbweb.nih.gov/ij/

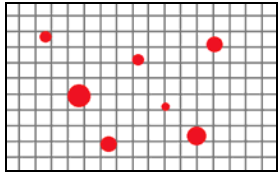 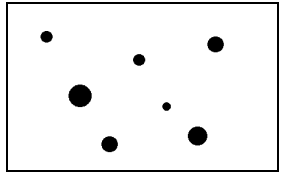 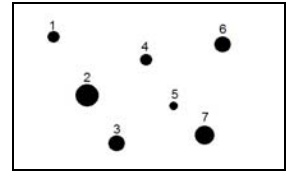

**Figure 2.** Schematic image of filter with blood clots.

**Figure 3.** Image after applying a binary filter.

**Figure 4.** Image after connected component labeling. Each clot is detected as a separate component.

## 2. Discussion and Concluding Remarks

In a clinical setting a manifold of at least two branches including a blood filter can be assembled downstream in the extracorporeal circuit, e.g., a dialysis circuit, an ECMO or a cardiopulmonary bypass circuit. By alternating the flow direction through the other branches of the manifold the study branch will be excluded with no interruption of the flow in the circuit. The study branch will be flushed and irrigated with normal isotonic saline until all blood is cleared from the branch through the filter. At this point only trapped clots will remain on the white surface of the blood filter. A digital picture taken from this filter will reveal only the clots. Analysis of the image will provide a quantification of the clot generated in the system. The excluded branch can be placed back in the circuit for continued usage and periodic evaluation of the clot in the system.

Using this method, it is not necessary to interrupt the flow in the circulation for clot quantification. Moreover, the quantification could be performed at periodic, lapsed, or sequential intervals. An automated device using this system could be employed in any extracorporeal circulation with timed process of branch exclusion from flow, irrigation, image acquisition and analysis, and resuming flow in the branch. After calibration such a system would be able to detect the initiation of a clot formation process in a circuit or alternatively detect when the quantity of the clots is above the acceptable threshold and the risk of thrombosis is high.

The technique described in this paper provides a valuable method for quantification of clots in an extracorporeal circuit. It can be used both in clinical settings to assess the level of clot formation, or in research settings as a tool for evaluation and comparison of anti-thrombotic measures.